%% file: main.tex
\documentclass[letterpaper]{article} 
\usepackage{aaai2026}  
\usepackage{times}  
\usepackage{helvet}  
\usepackage{courier}  
\usepackage[hyphens]{url}  
\usepackage{graphicx} 
\usepackage{natbib}  
\usepackage{caption} 
\usepackage{algorithm}
\usepackage{algorithmic}

\usepackage{xcolor}
\usepackage{amsmath}
\usepackage{amsfonts}
\usepackage{subcaption}
\usepackage{multirow}

\usepackage{newfloat}
\usepackage{listings}
\DeclareCaptionStyle{ruled}{labelfont=normalfont,labelsep=colon,strut=off} 
\floatstyle{ruled}
\newfloat{listing}{tb}{lst}{}
\floatname{listing}{Listing}
\title{CIP-Net: Continual Interpretable Prototype-based Network}
\author {
    Federico Di Valerio\textsuperscript{\rm 1},
    Michela Proietti\textsuperscript{\rm 1},
    Alessio Ragno\textsuperscript{\rm 2,\rm 3},
    Roberto Capobianco\textsuperscript{\rm 4}
}
\affiliations {
    \textsuperscript{\rm 1}Department of Computer, Control and Management Engineering (DIAG), Sapienza University, IT-00185, Rome, Italy\\
    \textsuperscript{\rm 2}INSA Lyon, CNRS, LIRIS UMR 5205, FR-94276, Villeurbanne France\\
    \textsuperscript{\rm 3}EPITA Research Laboratory (LRE), FR-69621, Le Kremlin-Bicêtre, France\\
    \textsuperscript{\rm 4}Sony AI, CH-8952,  Zurich, Switzerland\\
    \{divalerio, mproietti\}@diag.uniroma1.it, 
    alessio.ragno@\{insa-lyon.fr, epita.fr\},
    roberto.capobianco@sony.com
}

\begin{document}

\maketitle

\begin{abstract}
Continual learning constrains models to learn new tasks over time without forgetting what they have already learned. A key challenge in this setting is catastrophic forgetting, where learning new information causes the model to lose its performance on previous tasks. Recently, explainable AI has been proposed as a promising way to better understand and reduce forgetting. In particular, self-explainable models are useful because they generate explanations during prediction, which can help preserve knowledge. However, most existing explainable approaches use post-hoc explanations or require additional memory for each new task, resulting in limited scalability. In this work, we introduce CIP-Net, an exemplar-free self-explainable prototype-based model designed for continual learning. CIP-Net avoids storing past examples and maintains a simple architecture, while still providing useful explanations and strong performance. We demonstrate that CIP-Net achieves state-of-the-art performances compared to previous exemplar-free and self-explainable methods in both task- and class-incremental settings, while bearing significantly lower memory-related overhead. This makes it a practical and interpretable solution for continual learning.
\end{abstract}

\begin{links}
    \link{Code \& Supplementary Material}{https://github.com/KRLGroup/CIP-Net}
\end{links}

\section{Introduction}

Continual learning (CL) research aims at developing models that can learn tasks sequentially, without revisiting past data. In this setting, the major obstacle is \emph{catastrophic forgetting} \citep{french1999catastrophic,goodfellow2013empirical,ewc}: learning new tasks overwrites or distorts earlier knowledge, sharply degrading performance on past tasks.
Explainable artificial intelligence (XAI) helps with understanding this phenomenon by analyzing why and how certain knowledge is retained or lost across tasks. In this line of work, \citet{cossu2023protocol} leverage XAI to study explanation drift, i.e., a shift in the explanations of past tasks as new knowledge is acquired. Given that these representational modifications are usually a direct cause of catastrophic forgetting, several works directly exploit explanations to develop XAI-guided approaches that prevent explanation drift and, in turn, mitigate forgetting. Most methods use post-hoc explainability \citep{rnf,rrr,epr,samc} (i.e., explanations are produced after each task's training, looking at the model's output and parameters), whereas recent work introduces self-interpretable CL models \citep{icicle,proiettiprotocrl}. Among them, ICICLE \citep{icicle} builds on a popular prototype-based architecture, namely ProtoPNet \citep{protopnet}, to tackle catastrophic forgetting in continual image classification. Prototype-based networks directly build interpretability into the model, generating explanations during prediction \citep{rudin2019stop}. Specifically, these models link each prediction to a handful of visually meaningful image patches, thus enabling intuitive \emph{this-looks-like-that} reasoning that is self-generated rather than post-produced \citep{protopnet}. In CL, this explicit evidence makes drift easier to detect and track. However, ICICLE comes with one major drawback: it adds a new prototype head for each task, where a prototype head is a specific module that enables prototype-based reasoning in the model. As a consequence, computation and memory grow with the number of tasks, limiting ICICLE's scalability, and separate heads limit cross‑task sharing.

To overcome these issues, we propose a novel Continual Interpretable Prototype-based Network (CIP-Net), built on PIP-Net's \citep{pipnet} lighter prototype-based reasoning. A single fixed-size pool of shared patch-level prototypes yields robust accuracy and stable explanations, eliminating the need for exemplar storage or multiple task-specific prototype layers. To prevent interference in the shared prototype layer, we adopt alignment–uniformity self-supervision and targeted regularization mechanisms. On CUB-200-2011 (CUB, \citet{cubdataset}) and Stanford Cars (CARS, \citet{stanfordcars}) benchmarks, CIP-Net outperforms ICICLE and other exemplar-free standard CL approaches in both \textit{task-incremental learning} (TIL, task identity known at test time) and \textit{class-incremental learning} (CIL, task identity not known at test time). Our contributions are:
\begin{itemize}
    \item CIP-Net: an exemplar-free, self-explainable, prototype-based CL model with a shared prototype layer, that enables knowledge sharing and avoids the memory-related overhead associated with task-specific prototype heads;
    \item Introduction of targeted prototype regularization, discouraging changes in important prototypes for past tasks, and loss terms to encourage the use of sparse prototypes and promote orthogonality between classification heads;
    \item CIP-Net mitigates catastrophic forgetting, outperforming ICICLE in both TIL and CIL scenarios on CUB and CARS datasets, with gains up to $+35\%$ (TIL) and $+11.9\%$ (CIL) in final average accuracy on CUB and up to $+34.2\%$ (TIL) and $+38.2\%$ (CIL) on CARS;
    \item  Explanation drift analysis, which reveals that explanations remain stable during training;
    \item An open-source implementation of CIP-Net

\end{itemize}

\section{Related Work}
CL methods can be broadly framed within a five-branch taxonomy \citep{cl}. Regularization-based approaches introduce additional penalties to discourage changes in weights or activations. For instance, EWC \citep{ewc} estimates parameter importance via the Fisher information matrix, penalizing changes to high-importance weights. In contrast, LwF \citep{lwf} distills the previous model’s outputs on new-task data to preserve old decision boundaries. Differently, replay-based strategies approximate past data by storing a small exemplar buffer \citep{icarl,lucir} or generating pseudo-samples \citep{shin2017continual}, which are then replayed alongside the current task's data. Representation-based methods are typically exemplar-free alternatives, producing robust representations through self-supervised learning, large-scale pre-training, or representation learning. Examples belonging to this category are FeTrIL \citep{petit2023fetril}, which learns a feature-translation module to align new representations with a frozen old classifier, and PASS \citep{pass}, which combines prototype augmentation with self-supervision to maintain class discrimination without storing images. 
Optimization-based approaches, on the other hand, directly act on the optimization process, by projecting gradients in non-interfering directions \citep{ogd}, or using meta-learning \citep{itaml}. Finally, architecture-based solutions either allocate task-specific parameters through masking \citep{mallya2018piggyback}, pruning-and-packing \citep{mallya2018packnet}, or dynamic expansion \citep{yan2021dynamically}, to reduce task interference.
Within these families, some methods effectively leverage XAI. For example, LwM \citep{lwm} extends the idea behind LwF by distilling attention/activation maps instead of (or in addition to) logits, enabling exemplar-free knowledge transfer. Others employ explanations to select the examples to store in the replay buffer \citep{epr,aser,samc}. \citet{rnf}, instead, propose to create task-specific network components.  In these contexts, XAI allows for the identification of a phenomenon known as explanation drift, which consists of a change in the explanations of old tasks after training on subsequent ones \citep{explanation-drift}. By attenuating explanation drift, XAI-guided approaches prove to be effective in mitigating catastrophic forgetting \citep{xaiguidedcl, rrr}.

While most of the existing XAI-guided CL approaches are post-hoc, meaning they provide explanations for pre-trained models, ICICLE \citep{icicle} builds on ProtoPNet \citep{protopnet}, a prototype-based self-interpretable architecture. This type of architectures learn a set of prototypes that represent parts of the training input images, and perform the classification based on the similarity of the patches of the current input to the learned prototypes. While this scheme allows providing explanations at prediction time, ICICLE shows that it can also be exploited for suppressing catastrophic forgetting. To adapt this architecture to the CL scenarios, ICICLE features task-specific prototype layers and classification heads. However, this means that the network grows considerably with the number of tasks.

In this work, we take inspiration from this idea and specifically address these issues. We propose CIP-Net, which relies on a different prototype-based network, namely PIP-Net \citep{pipnet}. While replicating ICICLE’s exemplar-free protocol, we replace multiple prototype heads with a \emph{single and shared} prototype layer.
In this way, we manage to keep the memory and computational overhead introduced by prototype-based reasoning constant as the number of tasks grows, while maintaining interpretability and improving accuracy.

\section{Background}

In this section, we introduce the mathematical basis necessary to understand our proposed method. We begin by introducing prototype-based self-explainable models, which offer an interpretable alternative to standard deep learning approaches by providing interpretable predictions through comparisons of input features to learned prototypical parts. In particular, we focus on ProtoPNet \citep{protopnet} and then PIP-Net \citep{pipnet}, which are at the basis of ICICLE and our proposed approach, CIP-Net.

\subsection{Prototype-based Neural Networks}
A prototype-based neural network is an architecture that consists of a feature extractor followed by a prototype layer \( g_{p} \), which learns a set of \( P \) prototypical representations used for classification. Formally, a feature extractor \( f \) maps an input \( \mathbf{x} \in \mathbb{R}^{H \times W \times C} \) to a latent representation \( \mathbf{z} = f(x) \in \mathbb{R}^{H' \times W' \times D} \), and the prototype layer $g_p$ computes class evidence by comparing this latent representation to a set of prototypes using a model-specific similarity function.

In ProtoPNet \citep{protopnet}, each prototype \( \mathbf{p}_j \in \mathbb{R}^{H_p \times W_p \times D} \) is a learnable parameter tensor representing a prototypical latent patch. Similarity is measured using the negative squared Euclidean distance:
\begin{equation}
\text{sim}(\mathbf{z}_i, \mathbf{p}_j) = -\|\mathbf{z}_i - \mathbf{p}_j\|_2^2,
\end{equation}
where \( \mathbf{z}_i \) is a patch from the latent representation. For each prototype, the maximum similarity across all patches is used. The final class logit for class \( k \) is computed as:
\begin{equation}
\mathbf{s}_k = \sum_{j=1}^P w_{j,k} \cdot \max_i \text{sim}(\mathbf{z}_i, \mathbf{p}_j),
\end{equation}
where \( w_{j,k} \) is the learned importance weight of prototype \( \mathbf{p}_j \) for class \( k \). Additionally, during training, each prototype is projected onto the closest latent patch in the training set to ground it in a real image. The training loss combines cross-entropy and interpretability terms:
\begin{equation}
\mathcal{L}_{\text{ProtoPNet}} = \mathcal{L}_{\text{CE}} + \lambda_{\text{clst}} \cdot \mathcal{L}_{\text{clst}} + \lambda_{\text{sep}} \cdot \mathcal{L}_{\text{sep}},
\end{equation}
where \( \mathcal{L}_{\text{clst}} \) pulls latent patches closer to their class prototypes, and \( \mathcal{L}_{\text{sep}} \) pushes them away from other-class prototypes.

In PIP-Net \citep{pipnet}, prototypes are not learned directly but they are simply one channel of the final feature map. The presence of a prototype in the latent representation is determined through a softmax operation across the channel dimension. Given a latent feature map \( \mathbf{z} \in \mathbb{R}^{H' \times W' \times D} \), a softmax is applied at each spatial location:
\begin{equation}
z'_{h,w,d} = \frac{\exp(z_{h,w,d})}{\sum_{d'=1}^D \exp(z_{h,w,d'})},
\end{equation}
and the presence score of prototype \( d \) is obtained via spatial max-pooling:
\begin{equation}
p_d = \max_{h,w} z'_{h,w,d}.
\end{equation}
Classification scores are computed using a sparse, non-negative linear layer:
\begin{equation}
s_k = \sum_{d=1}^D w_{d,k} \cdot p_d,
\quad \text{with } w_{d,k} \geq 0.
\end{equation}

Due to the lack of specific prototype representations and the impossibility of performing the prototype projection, in order to ensure semantic consistency, PIP-Net employs a contrastive learning objective during pretraining. The full loss is:
\begin{equation}
\mathcal{L}_{\text{PIP-Net}} = \lambda_C \mathcal{L}_{\text{CE}} + \lambda_A \mathcal{L}_{\text{A}} + \lambda_T \mathcal{L}_{\text{T}},
\end{equation}
where \( \mathcal{L}_{\text{A}} \) aligns prototype activations between augmented views of the same image:
\begin{equation}
\mathcal{L}_{\text{A}} = -\frac{1}{HW} \sum_{h,w} \log \left( \mathbf{z}^{\prime}_{h,w,:} \cdot \mathbf{z}^{\prime\prime}_{h,w,:} \right),
\end{equation}
and \( \mathcal{L}_{\text{T}} \) encourages uniform prototype usage:

\begin{equation}
\mathcal{L}_{\text{T}} = -\frac{1}{D} \sum_{d=1}^D \log\left( \tanh \left( \sum_{b=1}^B \mathbf{p}^{(b)} \right) + \epsilon \right).
\end{equation}
Here, \( \mathbf{z}^{\prime} \) and \( \mathbf{z}^{\prime\prime} \) are feature maps from two augmentations of the same image, and \( \mathbf{p}^{(b)} \) is the prototype presence score vector in minibatch \( b \). This design allows PIP-Net to produce interpretable predictions using shared, semantically meaningful prototypes.

\subsection{ICICLE} 
\label{sec:icicle}
ICICLE \citep{icicle} extends prototype–based reasoning to exemplar-free CL. Assume a stream of $T$ tasks $\bigl(C^1,X^1,Y^1,\bigr),\dots,\bigl(C^T,X^T,Y^T\bigr)$, where $C^t$ is the set of classes introduced at task $t$, $X^t$ and $Y^t$ the associated samples and respective labels. During task $t$, only $X^t, Y^t$ are accessible, and no data from earlier tasks is stored. Clearly, \(C^i \cap C^j = \emptyset \quad \text{for all } i \neq j\).

While ICICLE is built on top of ProtoPNet, the authors propose to use a dedicated prototype layer $g^t$ and classification head $h^t$ for every task. Each prototype layer $g^t$ contains $M^t=K\cdot |C^t|$ prototypes, with $K$ being the number of prototypical parts per class. 

During training, $\mathcal{L}_{CE}$ is calculated on the concatenation of all task logits. On the other hand, $\mathcal{L}_{clst}$ and $\mathcal{L}_{sep}$ are left unaltered and computed within the $g^t$ head. Inspired by \citet{keswani2022proto2proto}, a regularization term prevents explanation shift by minimizing the difference between prototype similarities at tasks $t$ and $t-1$:
\begin{equation}
\mathcal{L}_{\text{IR}}=
\sum_{i=1}^{H}\sum_{j=1}^{W}
\bigl|
\operatorname{sim}\bigl(p^{\,t-1},z_{i,j}^{\,t}\bigr)-
\operatorname{sim}\bigl(p^{\,t},z_{i,j}^{\,t}\bigr)
\bigr|\,S_{i,j},
\label{eq:LIR}
\end{equation}
where $p^{t-1}$ is a prototype frozen after task $t-1$, $p^t$ its current counterpart, and $S$ is a binary mask selecting the $\gamma$-quantile of the pixels with the highest similarity.

\begin{figure*}
\centering
    \includegraphics[width=0.9\textwidth]{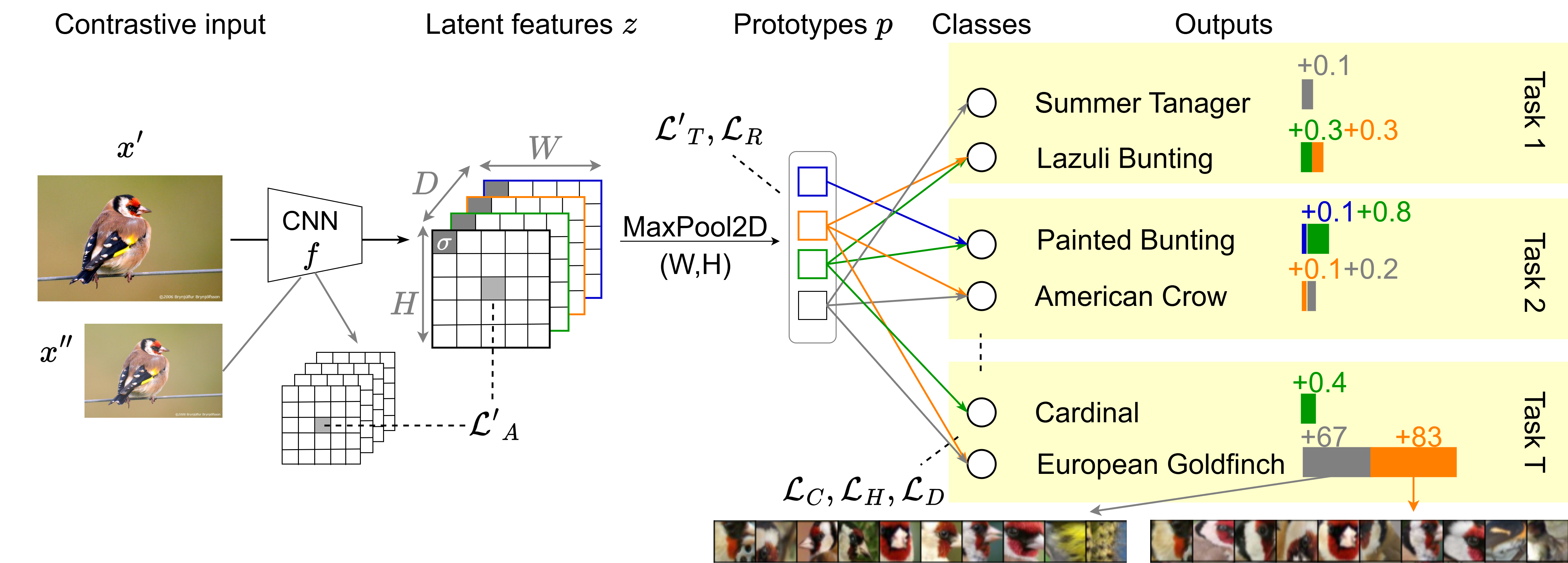}
    \caption{
    CIP-Net overview. Two augmented views are processed by a CNN to produce a prototype feature map $\mathbf{z}$; channel-wise softmax and maxpooling operation yield prototype-presence scores $\mathbf{p}$. Contrastive alignment loss $\mathcal{L}'_{A}$ aligns prototypes across views, $\mathcal{L}'_{T}$ promotes rarely used prototypes, and $\mathcal{L}_{R}$ limits drift of important ones over tasks. Task-specific sparse, non-negative heads map prototypes to classes, trained with negative log-likelihood loss $\mathcal{L}_{C}$ plus sparsity $\mathcal{L}_{H}$ and head-decorrelation $\mathcal{L}_{D}$.
}
    \label{fig:cipnet}
\end{figure*}

\section{Methods}
\label{sec:methods}
We propose CIP-Net, a CL model that combines interpretability, efficiency, and robustness to catastrophic forgetting. CIP-Net, illustrated in Figure \ref{fig:cipnet}, extends PIP-Net by adapting the shared prototype layer and introducing targeted task-aware regularization to incremental learning scenarios. In this section, we describe the architectural modifications, prototype pretraining, and full training objective.

\paragraph{Architecture}
CIP-Net builds upon the prototype extraction pipeline of PIP-Net: each input image is processed by a convolutional backbone that produces a latent feature map; prototype presence scores are computed via a softmax over the channel dimension, followed by spatial max-pooling, yielding a vector $\mathbf{p} = [p_1, ..., p_D] \in \mathbb{R}^D$. A known issue in CL arises when logits from newer tasks exhibit higher magnitudes than those from earlier ones. This scale mismatch introduces a bias toward recent tasks, thereby accelerating catastrophic forgetting. To address this, we modify the classification head to normalize both the prototype activation vector $\mathbf{p}$ and the classifier weight vectors $\mathbf{w}_{c}^{(t)} \in \mathbb{R}^D$ for each class $c$ in task $t$. The class scores are then computed as:
\begin{equation}
s_k = \tau \cdot \frac{\mathbf{p}^{\top} \mathbf{w}^{(t)}_c}{\|\mathbf{p}\|_2 \cdot \|\mathbf{w}^{(t)}_c\|_2}
\end{equation}
where $\tau$ is a learnable temperature parameter that compensates for the reduced variance introduced by normalization, ensuring that logits remain expressive and comparable across tasks.

\paragraph{Training}

Following the structure of PIP-Net, for each task, our training is composed of two distinct phases: a pretraining phase aimed at producing semantically rich and diverse prototypes, and a continual training phase in which task-specific classifiers are learned.

In the first phase, the convolutional backbone and prototype layer are trained jointly using an unsupervised objective that does not rely on class labels. The goal here is to shape the prototypes into meaningful and reusable concepts before introducing the task-specific classification.  
From PIP-Net, we modify slightly \(\mathcal{L}_A\) to encourage faster convergence between the two augmented views to their midpoint representation: 
\begin{equation}
\mathcal{L'}_{\text{A}} = -\frac{1}{2HW} \sum_{h,w} \log \left( (\mathbf{z}^{\prime}_{h,w,:} \cdot \mathbf{z}^{m}_{h,w,:})(\mathbf{z}^{\prime\prime}_{h,w,:} \cdot \mathbf{z}^{m}_{h,w,:}) \right),
\end{equation}
where \(\mathbf{z}^{m}_{h,w,:} = \frac{\mathbf{z}^{\prime}_{h,w,:} + \mathbf{z}^{\prime\prime}_{h,w,:}}{2}\).
Given that a trivial solution to \( \mathcal{L'}_A \) would be to collapse onto a single active prototype, we adopt the same solution as PIP-Net with a diversity-promoting term 
\( \mathcal{L'}_{T} \).
We modify the diversity loss with a selection mechanism that filters the prototypes on which we want to induce uniform activation. The resulting loss is then:
\begin{equation}
\mathcal{L'}_{T} = - \frac{1}{|\mathcal{\tilde I}|}\sum_{d \in \mathcal{\tilde I}} \log\!\Bigl(\tanh\bigl(\sum_{b}^{B} \mathbf{p}_{b}\bigr)+\varepsilon\Bigr), \ \ \  \mathcal{\tilde I}=\bigcup_{t=1}^{\hat{t}} \mathcal{\tilde I}^{t}.
\end{equation}
Here $\hat{t}$ is the current task and \(\mathcal{\tilde I}\) is the set of rarely activated prototypes. 
To obtain \(\mathcal{\tilde I}\), for each head we select the least frequently activated prototypes ($< 75$th percentile of prototypes with a presence score $\geq 0.5$) and we consider them as \textit{rarely-used}. Essentially, \(\mathcal{L'}_{T}\) forces rarely-used prototypes to activate at least once per batch.
Additionally, we introduce another regularization term, which is responsible for controlling the explanation drift, that is, the change of prototype activations when learning successive tasks. The resulting prototype-stability regularizer acts separately on every previously learned classification head. In this case, for each task $t$, only \textit{highly-used} prototypes belonging to \(\mathcal{I}^{t} = P \setminus \mathcal{\tilde I} ^{t} \) enter the loss: 
\begin{equation}
\mathcal{L}_{R} \;=\; \sum^{t<\hat t}_{t=1} \mathcal{L}^{t}_{R} = \sum^{t<\hat t}_{t=1}\frac{1}{|\mathcal I^{t}|} \sum_{d\in\mathcal I^{t}} \bigl(\max_c |w^{t}_{d,c}|\bigr)\, \lVert f^{\hat{t}}_d - f^t_d\rVert_2,
\end{equation}
where $w^{t}_{c, d}$ are the class-prototype weights for the previous task $t$. In $\mathcal{L}^{t}_{R}$, the Euclidean distance between the current prototype feature vector $f_d$ and the prototype features obtained from the frozen model trained on the previous task $f^t_d$ is penalized and weighted by the prototype's maximum outgoing class weight. 
This term stops the semantics of prototypes that are critical for earlier tasks from drifting too much while leaving rarely activated prototypes free to adapt. As a result, $\mathcal{L_R}$ safeguards past performance and explanations without hindering the network's capacity to learn and update prototypes for the current task. Note that this loss term is only active in tasks following the first one.
The total pretraining objective is given by:
\begin{equation}
\mathcal{L}_{\text{pre}} = \lambda_A \mathcal{L'}_A + \lambda_T \mathcal{L'}_{T} + \lambda_R \mathcal{L}_R,
\end{equation}
where each term contributes to building a diverse and interpretable prototype set.

Once pretraining is complete, the model enters the incremental learning phase. For each new task \( \hat{t} \), only the classification head \( \mathbf{w}^{\hat{t}} \) is newly initialized and trained, while the backbone and prototype layer are fine-tuned and earlier heads are frozen. The core of the objective is the standard negative log-likelihood classification loss \( \mathcal{L}_C \), which ensures that the model correctly predicts the labels of the current task.
However, to maintain interpretability and prevent interference across tasks, we introduce two additional regularizers. The first is a Hoyer sparsity loss \citep{hoyer}:
\begin{equation}
    \mathcal{L}_{H} = \left( 1 - \frac{1}{|C^{\hat{t}}|} \sum_{c \in C^{\hat{t}}} \frac{ \sqrt{D} - \dfrac{\|\mathbf{w}^{\hat{t}}_{c}\|_1}{\|\mathbf{w}^{\hat{t}}_{c}\|_2 + \epsilon} }{ \sqrt{D} - 1 } \right).
\end{equation}
\(\mathcal{L}_H\) encourages each class in the current task to rely on a small number of prototypes. This leads to concise and disentangled explanations while limiting overlapping prototype usage. The second term is a head decorrelation loss:
\begin{equation}
    \mathcal{L}_{D} = \sum_{t=0}^{\hat t} \Bigl(\lVert \mathbf{S}^{t}\rVert -\lVert\operatorname{diag}\!\bigl(\mathbf{S}^{t}\bigr)\rVert\Bigr),
\end{equation}
where \(\mathbf{S}^{t}=\bigl({\mathbf{W}}^{(t)}\,{\mathbf{W}}^{(\hat{t})\top}\bigr)^{2}\) is the squared pairwise dot product between the weight vectors of the $t$-th head and the current head $\hat{t}$.
\(\mathcal{L}_D\) promotes orthogonality between the current classification head and those from previous tasks. By minimizing the off-diagonal elements of the dot-product matrices between heads, this term prevents new tasks from reusing prototype combinations already assigned to earlier classes, reducing representational interference and forgetting.
The full training objective for the current task is therefore:
\begin{equation}
\mathcal{L} = \mathcal{L}_{\text{pre}} + \lambda_C \mathcal{L}_C + \lambda_H \mathcal{L}_H + \lambda_D \mathcal{L}_D,
\end{equation}
where each component plays a distinct role: preserving diversity and stability from the pretraining phase, ensuring accurate task-specific predictions, and enforcing structured, sparse, and disjoint prototype usage over time.

\begin{table*}[t]
\centering
\small
\setlength{\tabcolsep}{3pt}
\begin{tabular}{|c|c|c|c|c|c|c|}
\hline
\textbf{Method} & \multicolumn{3}{c|}{\textbf{Final Avg. Task-Incremental Accuracy}} & \multicolumn{3}{c|}{\textbf{Final Avg. Class-Incremental Accuracy}} \\
\cline{2-7}
& 4 Tasks & 10 Tasks & 20 Tasks & 4 Tasks & 10 Tasks & 20 Tasks \\
\hline
EWC          & 0.445 $\pm$ 0.012 & 0.288 $\pm$ 0.034 & 0.188 $\pm$ 0.031 & 0.213 $\pm$ 0.007 & 0.095 $\pm$ 0.007 & 0.046 $\pm$ 0.011 \\
LWM          & 0.452 $\pm$ 0.023 & 0.294 $\pm$ 0.032 & 0.226 $\pm$ 0.025 & 0.180 $\pm$ 0.011 & 0.090 $\pm$ 0.011 & 0.044 $\pm$ 0.008 \\
LWF          & 0.301 $\pm$ 0.048 & 0.175 $\pm$ 0.028 & 0.129 $\pm$ 0.023 & 0.219 $\pm$ 0.011 & 0.078 $\pm$ 0.008 & 0.072 $\pm$ 0.008 \\
ICICLE      & \underline{0.654 $\pm$ 0.011} & \underline{0.602 $\pm$ 0.035} & \underline{0.497 $\pm$ 0.099} & \underline{0.350 $\pm$ 0.053} & \underline{0.185 $\pm$ 0.005} & \underline{0.099 $\pm$ 0.003} \\
\textbf{CIP-Net}      & \textbf{0.772 $\pm$ 0.015} & \textbf{0.846 $\pm$ 0.024} & \textbf{0.847 $\pm$ 0.016} & \textbf{0.469 $\pm$ 0.012} & \textbf{0.358 $\pm$ 0.030} & \textbf{0.180 $\pm$ 0.012} \\
\hline
FeTrIL & 0.750 $\pm$ 0.008 & 0.607 $\pm$ 0.018 & 0.407 $\pm$ 0.051 & 0.375 $\pm$ 0.006 & 0.199 $\pm$ 0.003 & 0.127 $\pm$ 0.011 \\
PASS    & 0.775 $\pm$ 0.006 & 0.647 $\pm$ 0.006 & 0.518 $\pm$ 0.012 & 0.395 $\pm$ 0.001 & 0.233 $\pm$ 0.009 & 0.139 $\pm$ 0.017 \\
\hline
PIP-Net C   & \multicolumn{6}{c|}{0.843 $\pm$ 0.002} \\
\hline
\end{tabular}
\caption{Final average (3 seeds) accuracy comparison for different numbers of tasks on CUB for both task- and class-incremental scenarios. This table describes the behavior of CIP-Net with an increasing number of tasks to be learned and compares it to other baselines. CIP-Net outperforms the baseline methods across all task numbers. Additionally, we show the gap between interpretable and black-box models by
comparing CIP-Net to FeTrIL \citep{petit2023fetril} and PASS \citep{pass}. At last, the PIP-Net (with ConvNeXt \citep{convnexttiny} backbone) performance upper bound is presented.
Results of methods different from CIP-Net are taken from \citet{icicle}.}
\label{table:avg_acc}
\end{table*}

\section{Results}

We evaluate CIP-Net on the standard CL benchmarks in the prototypes literature of CUB \citep{cubdataset} (200 bird species) and CARS \citep{stanfordcars} (196 car models). We split the classes in each dataset into \{4, 10, 20\} and \{4, 7, 14\} tasks, respectively. We run each experiment on 3 different seeds, except for the ablation study (1 seed).
Further details on compute resources and used hyperparameter values are described in the \textit{Supplementary Material}.

\begin{figure}[h]
    \centering
    \begin{subfigure}[b]{0.2325\textwidth}
        \centering
        \includegraphics[width=\textwidth]{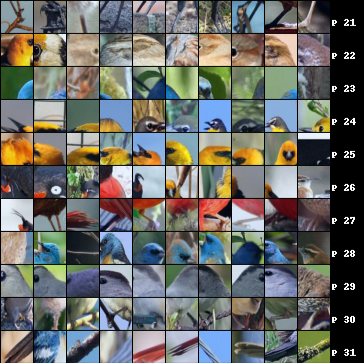}
        \caption{Task 0}
        \label{fig:1}
    \end{subfigure}
    \begin{subfigure}[b]{0.2325\textwidth}
        \centering
        \includegraphics[width=\textwidth]{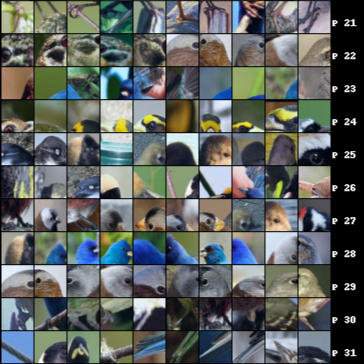}
        \caption{Task 1}
        \label{fig:2}
    \end{subfigure}
    \begin{subfigure}[b]{0.2325\textwidth}
        \centering
        \includegraphics[width=\textwidth]{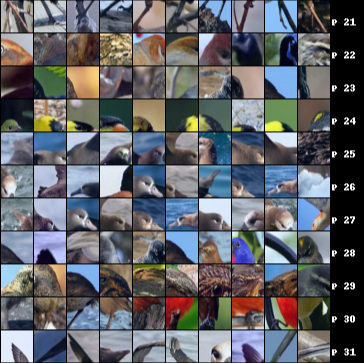}
        \caption{Task 2}
        \label{fig:3}
    \end{subfigure}
    \begin{subfigure}[b]{0.2325\textwidth}
        \centering
        \includegraphics[width=\textwidth]{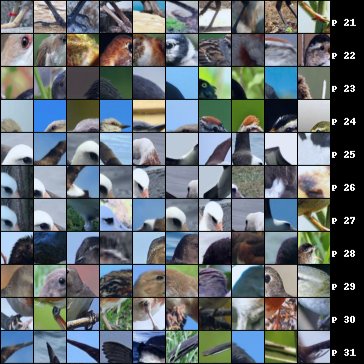}
        \caption{Task 3}
        \label{fig:4}
    \end{subfigure}

    \caption{Visualization of a portion of prototypes for each task in CUB's 4-task setting), obtained after training on the whole task sequence.}
    \label{fig:prototypes_pertaskaware}
\end{figure}

\paragraph{CIP-Net Outperforms Baselines in All Tested Scenarios}
We compare CIP-Net with ICICLE \citep{icicle} and the baselines it was compared against, as it is the current state-of-the-art exemplar-free interpretable CL method and, to our knowledge, the closest prototype-based CL approach to our method in the literature.
In Table \ref{table:avg_acc}, we report the final average accuracy, i.e., the average accuracy computed on all the tasks after training the model on the last one, obtained by the tested methods in the TIL and CIL scenarios on the CUB dataset. Final average accuracies for every task configuration on CARS and the per-task average accuracies for the 4-task setting on CUB (also with ResNet34 backbone) can be found in the \textit{Supplementary Material}.  We use these metrics as they are standard in CL literature \citep{icicle,cl}. CIP-Net consistently outperforms the baselines across all settings, with improvements over ICICLE in final average accuracy from $+11.5\%$ (4 tasks) to $+35.0\%$ (20 tasks) in TIL and from $+11.9\%$ (4 tasks) to $+8.1\%$ (20 tasks) in CIL on CUB. On CARS, we get even higher improvements ranging from $+20.8\%$ (4 tasks) to $+34.2\%$ (14 tasks) in TIL and from $+25.4\%$ (4 tasks) to $+38.2\%$ (7 tasks) and $+27.3\%$ (14 tasks) in CIL (see the \textit{Supplementary Material}).
Interestingly, in the TIL scenario, the performance remains high as the number of tasks increases, suggesting that CIP-Net better mitigates cumulative forgetting when the task identity is known. We hypothesize this could be the result of CIP-Net learning more generalizable prototypes, selectively re-adapting important ones, or progressively recognizing a broader set of patterns as new tasks are introduced,
while still treating specific prototypes as discriminative for earlier tasks, whose classification heads remain frozen. Some intuition of this can be noticed in Figure \ref{fig:prototypes_pertaskaware}, where the same prototypes are visualized for each task after training on the last one.
In contrast, a more typical trend appears in the CIL scenario. As the number of tasks increases, forgetting also increases, and inferring the correct class from the input alone becomes more challenging for the model. As it commonly happens with CL models in the CIL setting, CIP-Net also exhibits a bias toward the most recent task (shown for the 4-task settings on CUB in the \textit{Supplementary Material}), leading to reduced average accuracy as the task space expands.
However, CIP-Net still outperforms the considered baselines in all task settings, highlighting a better retention of past knowledge.

\begin{figure} 
    \centering
    \begin{subfigure}[b]{0.17\textwidth}
        \centering
        \includegraphics[width=\textwidth]{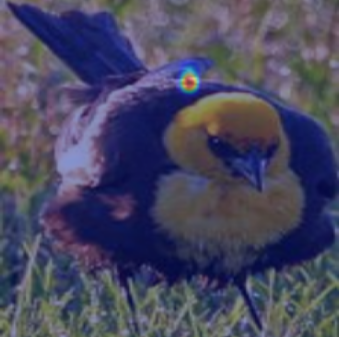}
        \caption{Task 1}
    \end{subfigure}
    \begin{subfigure}[b]{0.17\textwidth}
        \centering
        \includegraphics[width=\textwidth]{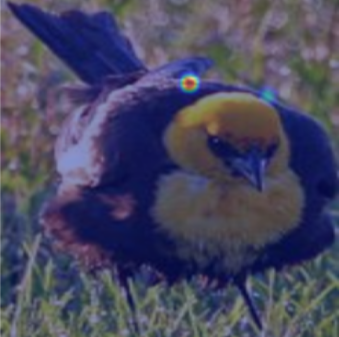}
        \caption{Task 2}
    \end{subfigure}

    \begin{subfigure}[b]{0.17\textwidth}
        \centering
        \includegraphics[width=\textwidth]{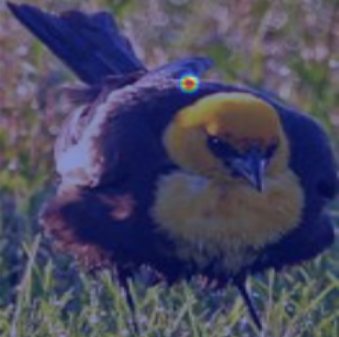}
        \caption{Task 3}
    \end{subfigure}
    \begin{subfigure}[b]{0.17\textwidth}
        \centering
        \includegraphics[width=\textwidth]{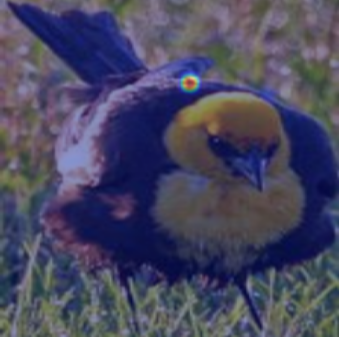}
        \caption{Task 3}
    \end{subfigure}

    \caption{Example of suppressed prototype activation drift for one of the regularized prototypes from task 1 to task 4 (4-tasks setting) on CUB. More details in Supplementary Material.}
    \label{fig:conceptdrift}
\end{figure}

\begin{figure}[h]
    \centering
    \begin{subfigure}[b]{0.23\textwidth}
        \centering
        \includegraphics[width=\textwidth]{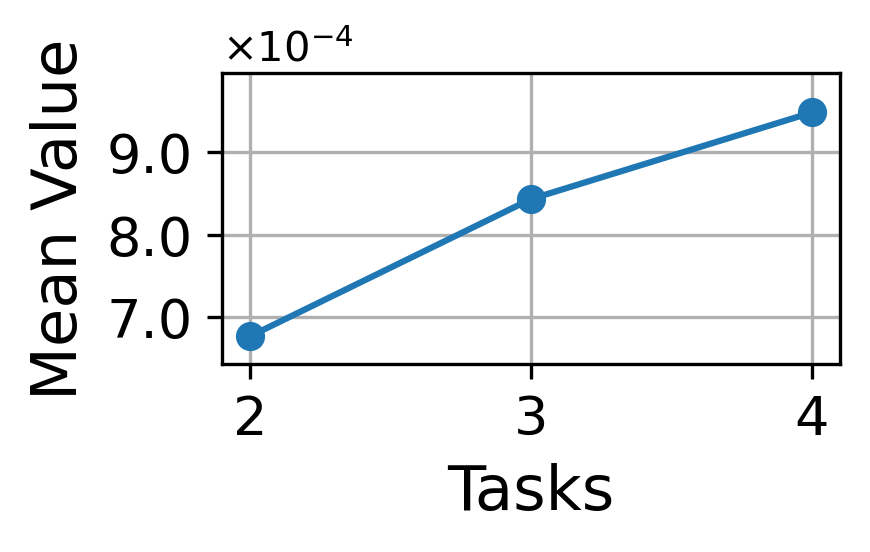}
        \caption{}
    \end{subfigure}    
    \begin{subfigure}[b]{0.23\textwidth}
        \centering
        \includegraphics[width=\textwidth]{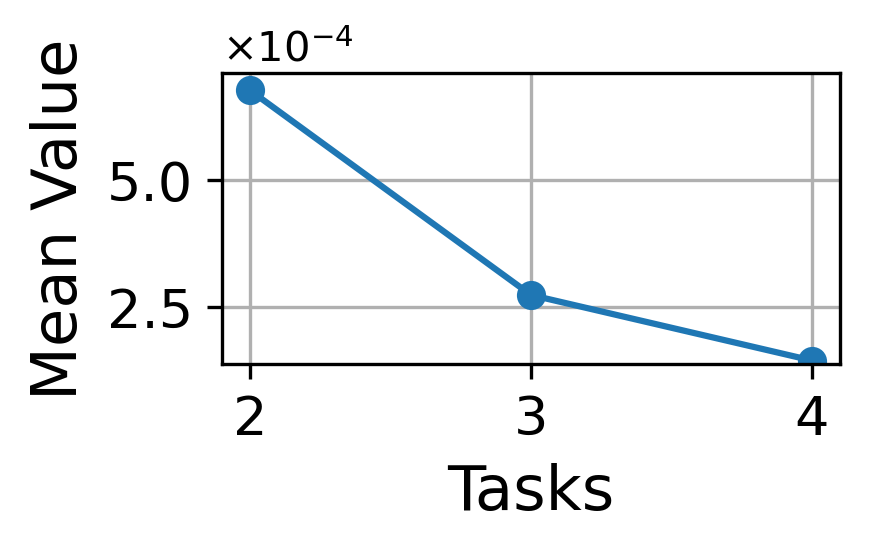}
        \caption{}
    \end{subfigure}
    \caption{(a) Average difference of activation values over classes between the current task and the first task. (b) Average difference of activation values over classes between the current task and the immediate previous task.}
    \label{fig:activations_line_mean}
\end{figure}

\paragraph{CIP-Net Mitigates Explanation Drift}
To assess whether our XAI-guided learning truly mitigates explanation drift, we monitor how the activations of prototypes learned in the earliest tasks on specific input images evolve as subsequent tasks are introduced. Figures \ref{fig:conceptdrift} shows how the activation of one of the regularized prototypes does not change from the first task to the last one (4-task setting). This example offers a qualitative confirmation that CIP-Net’s regularization term suppresses modifications to the most frequently used prototypes, thereby minimizing inter-task interference and keeping their activations essentially stable across the entire training process. Examples of other images together with a rare example of a drifted regularized prototype are shown in the \textit{Supplementary Material}. 
A quantitative evaluation of the suppression of the drift is presented in Figure \ref{fig:activations_line_mean}: panel (a) plots the average change in prototype activations computed over classes between the current task and the first task, while panel (b) depicts the average change between the current task and the immediately preceding task. The plots clearly show that throughout the tasks, the activations stay mostly stable.

\begin{table}[h]
\centering
\small
\setlength{\tabcolsep}{1pt}
\begin{tabular}{|l|c|c|c|c|c|}
\hline
\textbf{Model} & \textbf{\# Prototypes} & \textbf{Total Parameters (4-tasks)} \\
\hline
\textbf{ICICLE} & \( \sum^{T}_{t} M^t = \sum^{T}_{t} K \cdot |C_t|\) & \(\sum^{T}_{t} M^t \cdot \#(\mathbf{W}_{g_t}) \approx 1\text{MLN} \) \\
\hline
\textbf{CIP-Net} & \(\leq D\) & 0 \\
\hline
\end{tabular}
\caption{Comparison of parameters' overhead between CIP-Net and ICICLE used for the prototype layers. Note that \(\#(\mathbf{W}_{g_t})\) represents the number of parameters of layer \(g_t\) in ICICLE, where they use prototypes of dimension $512$.
Unlike ICICLE, CIP-Net does not use additional parameters for its prototype-based reasoning. The classification heads are not considered in the prototype layers.}
\label{tab:proto_params}
\end{table}

\paragraph{CIP-Net Reduces Overhead}
As described in the Methods section, CIP-Net integrates a single prototype layer as the final layer of the convolutional backbone. The number of prototypes, a predefined hyperparameter, serves as an upper bound since the model may use only a subset. This is a key difference from ICICLE, which introduces a separate prototypical part layer $g_t$ for each task, causing parameters to grow linearly with the number of tasks. Table \ref{tab:proto_params} compares the memory overhead from prototype-based reasoning in ICICLE and CIP-Net.
Instead of adding $M^t = K \cdot |C_t|$ new prototypes at each incremental step, CIP-Net reuses and incrementally updates a fixed, shared prototype pool, regularized to retain prior knowledge while adapting to new classes. This approach reduces prototype-based reasoning related storage and computation while preserving interpretability without ICICLE’s architectural growth. For each task, only a tensor of $D$ values is stored to record frequently used prototypes, enabling the regularization described in the Methods section.

\begin{table*}[h]
\centering
\small
\setlength{\tabcolsep}{1pt}
\begin{tabular}{|l|c|c|c|c|c|c|c|c|c|c|}
\cline{1-11}
& \multicolumn{5}{|c|}{\textbf{Task-Incremental Accuracy}} & \multicolumn{5}{|c|}{\textbf{Class-Incremental Accuracy}} \\
\cline{2-11}
\textbf{CIP-Net} & Task 1 & Task 2 & Task 3 & Task 4 & Final Avg. & Task 1 & Task 2 & Task 3 & Task 4 & Final Avg. \\
\hline
\textbf{w/o $\mathcal{L}_D$}      
& 0.72 $\pm$ 0.04 & 0.71 $\pm$ 0.13 & 0.81 $\pm$ 0.03 & 0.85 $\pm$ 0.05 & 0.77 $\pm$ 0.01 &  0.09 $\pm$ 0.01 & 0.29 $\pm$ 0.06 & 0.50 $\pm$ 0.01 & 0.82 $\pm$ 0.06 & 0.43 $\pm$ 0.00 \\
\textbf{w/o $\mathcal{L}_H$}      
& 0.85 $\pm$  0.01 & 0.72 $\pm$ 0.14 & 0.81 $\pm$ 0.03 & 0.84 $\pm$ 0.06 & 0.81 $\pm$ 0.03 & 0.19 $\pm$ 0.04 & 0.45 $\pm$ 0.05 & 0.61 $\pm$ 0.04 & 0.76 $\pm$ 0.07 & 0.50 $\pm$ 0.03\\
\textbf{w/o $\mathcal{L}_D \& \mathcal{L}_H$}      
& 0.81 $\pm$ 0.01 & 0.74 $\pm$ 0.13 & 0.82 $\pm$ 0.02 & 0.84 $\pm$ 0.05 & 0.80  $\pm$ 0.03 & 0.11 $\pm$ 0.03 & 0.24 $\pm$ 0.10 & 0.49 $\pm$ 0.04 & 0.82 $\pm$ 0.05 & 0.42 $\pm$ 0.03\\
\textbf{w/o $\mathcal{L}_R$}      
& 0.14 $\pm$ 0.02 & 0.35 $\pm$ 0.04 & 0.56 $\pm$ 0.04 & 0.86 $\pm$ 0.06 & 0.48 $\pm$ 0.03 & 0.00 $\pm$ 0.00 & 0.08 $\pm$ 0.05 & 0.17 $\pm$ 0.03 & 0.83 $\pm$ 0.07 & 0.27 $\pm$ 0.02 \\
\textbf{w/o $\tau$}      
& 0.83 $\pm$ 0.04 & 0.37 $\pm$ 0.04 & 0.32 $\pm$ 0.04 & 0.31 $\pm$ 0.08 & 0.46 $\pm$ 0.05 & 0.83 $\pm$ 0.04 & 0.02 $\pm$ 0.01 & 0.00 $\pm$ 0.00 & 0.00 $\pm$ 0.00 & 0.21 $\pm$ 0.01 \\
\hline
\textbf{CIP-Net}      
& 0.83 $\pm$ 0.00 & 0.63 $\pm$ 0.15 & 0.81 $\pm$ 0.04 & 0.82 $\pm$ 0.06 & 0.77 $\pm$ 0.01 & 0.25 $\pm$ 0.04 & 0.42 $\pm$ 0.02 & 0.51 $\pm$ 0.04 & 0.70 $\pm$ 0.06 & 0.47 $\pm$ 0.01 \\
\hline
\end{tabular}
\caption{Model's ablations (3 seeds) comparing accuracies in TIL and CIL 4-tasks scenarios.}
\label{table:ablations}
\end{table*}

\paragraph{Loss Components Functional Importance to CIP-Net's Performance}
To assess the importance of each loss term in maintaining CIP-Net’s performance and stability across tasks, we conduct an ablation study, with results reported in Table \ref{table:ablations}.

Removing the decorrelation loss $\mathcal{L}_D$ caused a slight TIL drop and a larger CIL decline, disproportionately affecting earlier tasks. Interestingly, omitting the Hoyer term $\mathcal{L}_H$ yielded performance comparable to the full model. Further analysis showed that setting to $0$ all near-zero weights below the $50$th percentile maintained accuracy, suggesting that classification heads naturally favor sparse weight matrices, enabling more compact explanations. This points to future work on promoting sparsity more efficiently or allowing it to emerge naturally.
When both $\mathcal{L}_D$ and $\mathcal{L}_H$ were removed, CIL accuracy declined, again affecting early tasks more. The absence of these regularizers loosens the constraints on the prototypes, which can reduce their intuitiveness and discriminative power and marginally lower overall performance. The prototype regularization term $\mathcal{L}_R$ proved essential: its removal caused one of the two worst performances in both CIL and TIL, with strong bias toward recent tasks and aggravated forgetting. Finally, excluding the temperature parameter $\tau$ produced a similar overall accuracy drop to removing $\mathcal{L}_R$, but with an opposite trend of better performance on the earliest task and severely reduced results on the most recent ones.

Overall, these results confirm that each component distinctly contributes to balancing stability, plasticity, and discriminative power in CIP-Net.

\paragraph{Global and Local Explanations}
Like PIP-Net, CIP‑Net also offers two complementary forms of interpretability. Global explanations are provided by the decision layers: the weights link each class to its relevant prototypes, clearly counting only those prototypes whose weights are non‑zero, with smaller values indicating less importance. 

Local explanations zoom in on a single prediction, pinpointing the prototypes that fire at specific spatial locations. This is valid for both TIL and CIL scenarios. Naturally, the particular subset of prototypes, and thus the visual rationale, may differ between TIL and CIL evaluations. 
An intuitive illustration of this patch-level rationale is shown in the \textit{Supplementary Material}.

To get a visual representation of each prototype $d$, we retrieve the top-$k$ training patches that maximize its importance score, computed as the element-wise product of the normalized prototype activations for sample $i$ and classification weights considered, based on the scenario being TIL or CIL: \(\mathbf{m}_d = \max_c \mathbf{p}^{\top}_i \mathbf{w}_d\).
We empirically found the image patches with a score $ \mathbf{m}_d> 0.01$ to be more intuitive, so only patches whose score exceeds this value are retained.
In the TIL case, the search is restricted to each task’s training images, so that for each task we get a full visual representation of the prototypes. Whereas in the CIL case, the full training set is considered.
An example for the $4$-tasks setting is shown in Figure \ref{fig:prototypes_pertaskaware}.

\section{Conclusions}

We have introduced CIP-Net, an exemplar-free, prototype-based CL framework that delivers state-of-the-art performance compared to previous exemplar-free and self-explainable methods in both TIL and CIL settings. 
By grounding every decision in a combination of shared prototypes, CIP-Net offers explanations that remain useful and stable across tasks, effectively mitigates catastrophic forgetting, and avoids the increasing memory-related overhead commonly associated with prototype-based reasoning.

\paragraph{Limitations}
CIP-Net retains the interpretability benefits of prototype-based models but also their drawbacks, including the latent–human semantic gap, vulnerability to small adversarial perturbations, and other known issues \citep{gautam2023looks,hoffmann2021looks,kim2022hive,nauta2021looks,rymarczyk2022interpretable}. Although the PIP-Net backbone generalizes beyond fine-grained datasets like CUB and CARS, CIP-Net has not yet been tested on them.
Our frequency-based regularization constrains only some prototypes, so an inappropriate proportion can blur explanations or reduce accuracy, particularly over long task sequences. Frequent prototypes are not always meaningful, as they may represent common backgrounds (e.g., sky or sea), though their usefulness can depend on context: for instance, in CUB, a “sea” prototype aids seabird classification.

\paragraph{Future Work}
Promising research directions include devising better sparsity-promoting strategies for more compact explanations and exploring alternative prototype regularization schemes to address the previously noted limitations.
Additionally, while PIP-Net is designed to recognize out-of-distribution samples, this has not been explored in CIP-Net yet; leveraging this capability for autonomous new-task detection could be an interesting direction.

\section{Acknowledgments}
This work was supported by the French National Research Agency (ANR) through the France 2030 program, under the PEPR “WAIT4” (ANR-22-PEAE-0008). This work has been carried out while Michela Proietti and Federico Di Valerio were enrolled in the Italian National Doctorate on Artificial Intelligence and the PhD in Engineering in Computer Science, respectively, run by Sapienza University of Rome.

\bibliography{aaai2026}
\newpage
\appendix
\section{Supplementary Material}
\input{appendix}

\end{document}

%% file: appendix.tex
\section{Additional Experimental Setup Details}
We used ConvNeXt-tiny \citep{convnexttiny} as backbone.
We performed a hyperparameter search for the learning rates, \(\lambda_R\),\(\lambda_H\). We used AdamW \citep{adamw} optimizers with the learning rates and respective settings shown in Table \ref{tab:hyperparams}.
We set a batch size of $64$ for all task settings and datasets apart for the $20$-task setting on the CUB-200-2011 dataset, where the batch size used was $32$. However, each mini-batch contains two views of an image, resulting in twice the dimension (batch sizes of 128 and 64).
Input image dimensions were $224 \times 224 \times 3$.
The backbone was pretrained on ImageNet and the classifiers' weights are initialized with values drawn from the normal distribution with mean $1.0$ and standard deviation $0.1$.
During pretraining, \(\lambda_A\) is slowly increased to 1 (to prevent the trivial solution that every patch gets the same encoding), \(\lambda_T = 1.0 , \lambda_R = 8.0\). During training \(\lambda_A = 5.0, \lambda_T = 1.0, \lambda_R = 8.0, \lambda_H = 10.0, \lambda_D = 0.005, \lambda_C = 2.0\).
For the first epoch of training of each task, backbone \(f\) is frozen and only weights \(\mathbf{W}^{\hat{t}}\) of the current classification head are trained. 
The predictions during training follow the CIL scenario: the output of each classification head is concatenated and passed to the final softmax to get the prediction.
During inference, prototype presence scores \(\mathbf{p}<0.1\) are ignored. 

Each model was trained in parallel on Tesla V100 GPUS.

\section{Task-specific Results on CUB}
In Table \ref{table:avg_acc_4ts} we show detailed average accuracies for each task in the 4-task setting on CUB-200-2011. CIP-Net outperforms all the baselines. A bias towards the last task can be noticed in the CIL scenario, as commonly happens to CL models. 

Our improvements mainly stem from reduced forgetting rather than backbone capacity. Additional preliminary experiments with ResNet34 as backbone also show that CIP-Net outperformed ICICLE in 4-task setting with the same hyperparameters used in the main experiments using ConvNeXt-tiny.

\section{Results on Stanford Cars}
Table \ref{table:avg_acc_cars} depicts how CIP-Net works on the Stanford Cars dataset compared to other baseline methods. Results are consistent with the ones on CUB and show that CIP-Net outperforms all the baselines considered.

\section{Prototype Activation Drift}
\begin{figure*}[h] 
    \centering
    \begin{subfigure}[b]{0.791\textwidth}
        \centering
        \includegraphics[width=\textwidth]{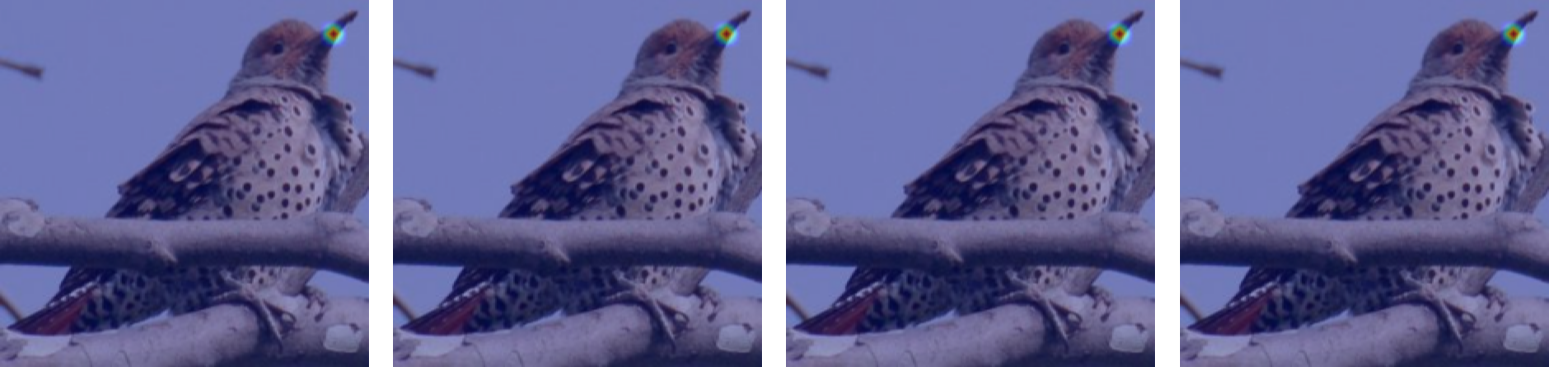}
    \end{subfigure}
    \begin{subfigure}[b]{0.8\textwidth}
        \centering
        \includegraphics[width=\textwidth]{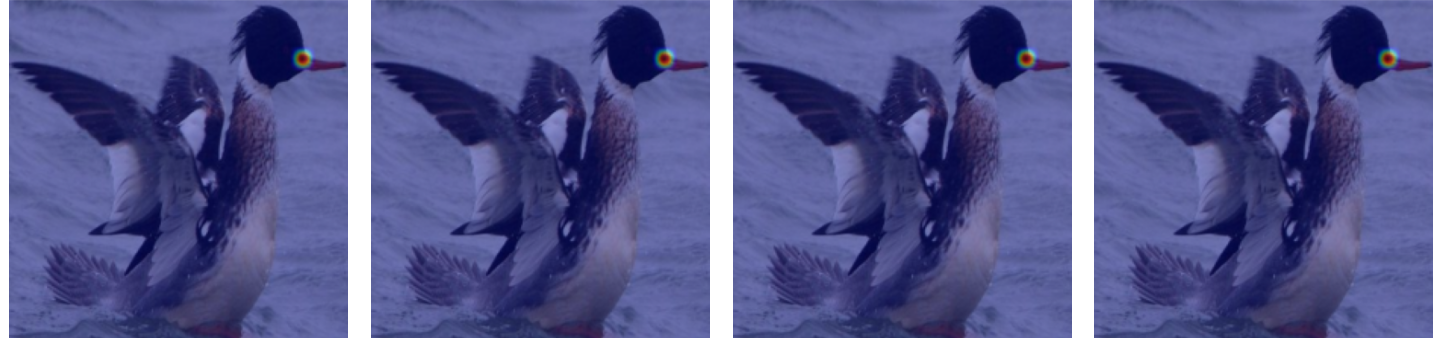}
    \end{subfigure}
    \begin{subfigure}[b]{0.791\textwidth}
        \centering
        \includegraphics[width=\textwidth]{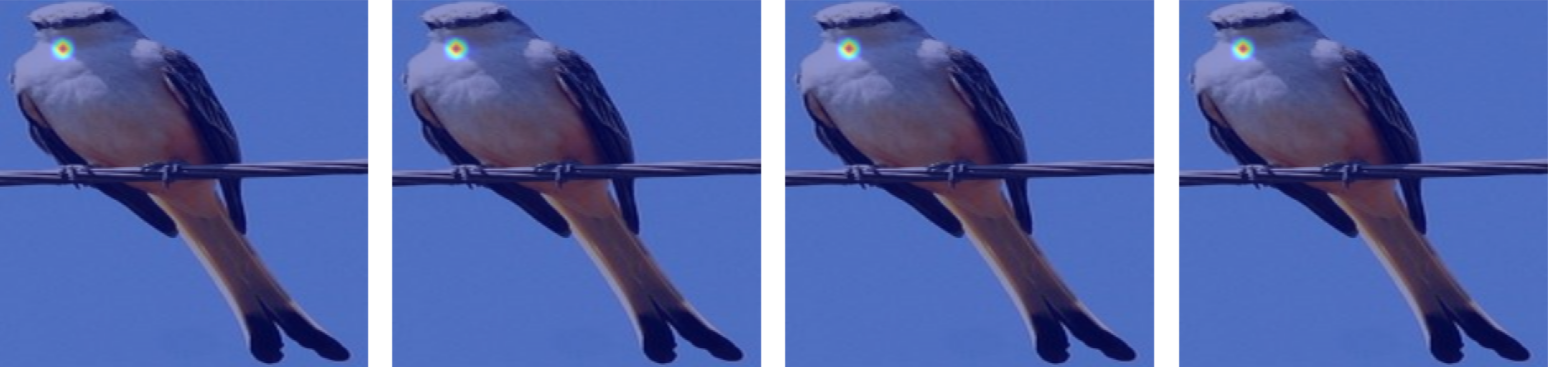}
    \end{subfigure}
    \begin{subfigure}[b]{0.8\textwidth}
        \centering
        \includegraphics[width=\textwidth]{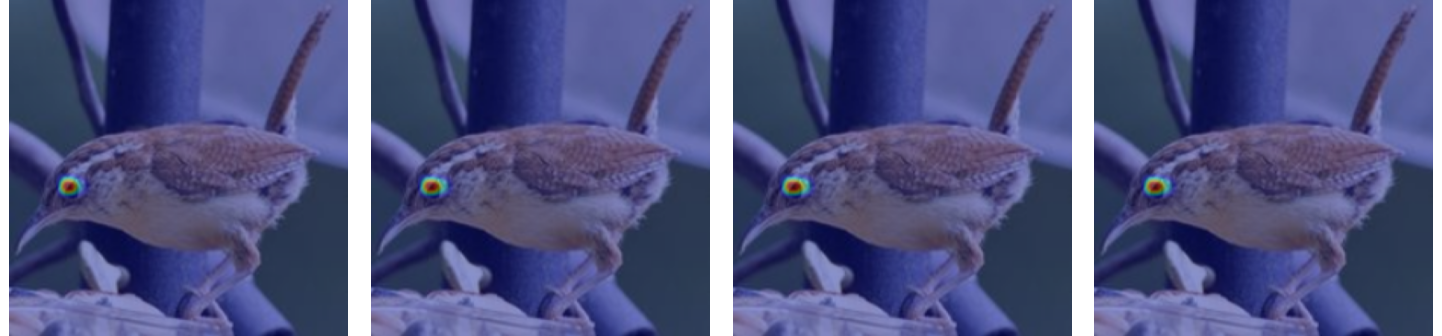}
        \caption{Example of suppressed prototype activation drift for some of the regularized prototypes from task 1 to task 4 (4-task setting) on CUB-200-2011.}
    \end{subfigure}
    \begin{subfigure}[b]{0.791\textwidth}
        \centering
        \includegraphics[width=\textwidth]{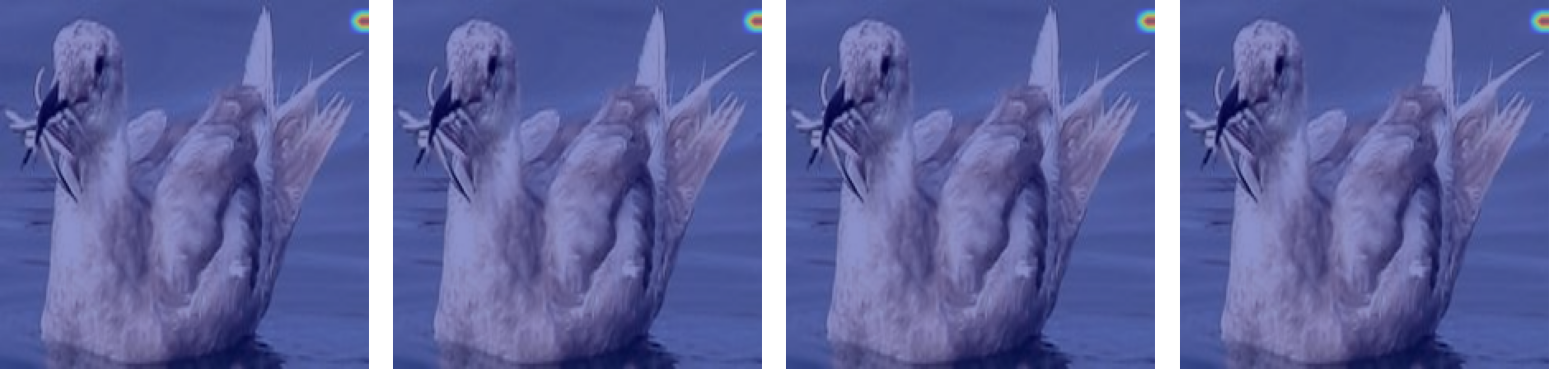}
        \caption{Example of "background" prototype.}
        \label{fig:background}
    \end{subfigure}
    \begin{subfigure}[b]{0.791\textwidth}
        \centering
        \includegraphics[width=\textwidth]{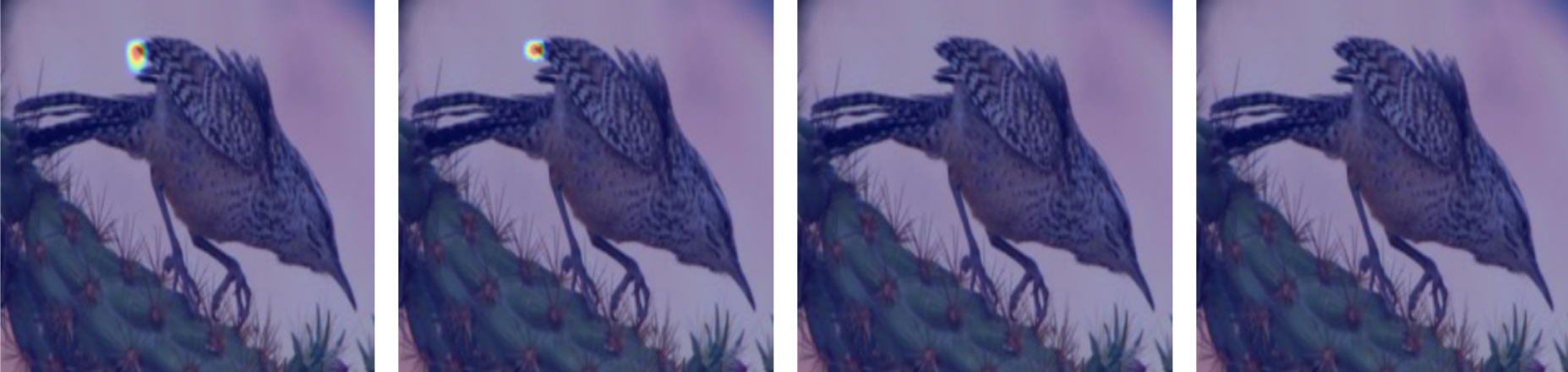}
        \caption{A rare case of prototype activation drift in a regularized prototype.}
        \label{fig:drift}
    \end{subfigure}

    \caption{
    (a) Examples of suppressed prototype activation drift for some of the regularized prototypes from task 1 to task 4 (4-task setting) on CUB-200-2011.
    (b) Example of prototype activation drift for a prototype that activated on the background of the input image. 
    (c) Rare example of prototype activation drift in a regularized prototype.}
    \label{fig:conceptdrift_supp}
\end{figure*}

In Figure \ref{fig:conceptdrift_supp}, we show other regularized prototypes that remain stable throughout training in the 4-task setting. While very rare, in Figure \ref{fig:drift}, we show a regularized prototype that drifted in the 4-task setting.
We also present an example of prototype activation drift in subfigure \ref{fig:background}, where a prototype was activated on a background region of the input image. Regarding this last case, it's important to emphasize that CIPNet’s prediction relies more on the combination of activated prototypes rather than the presence of any single one. As such, prototypes should be visualized collectively to fully understand an explanation.

\section{Local Explanations}
\begin{figure}[h]
    \includegraphics[width=0.48\textwidth]{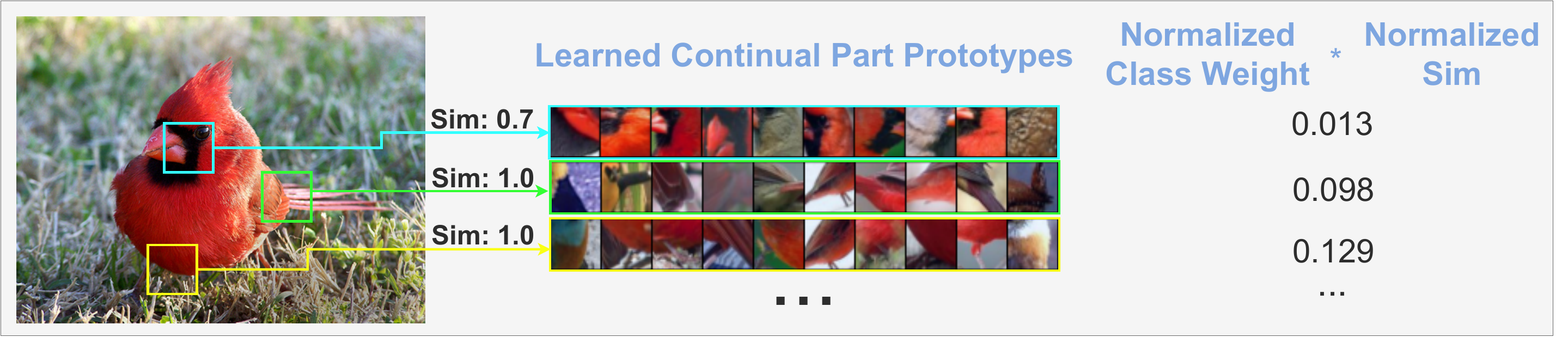}
    \caption{Example of local explanation of CIP-Net for the correct class. CIP-Net learns part-prototypes visualized as patches from the training data, and localizes similar image patches in an unseen input image.}
    \label{fig:predcipnet}
\end{figure}

For completeness in Figure \ref{fig:predcipnet}, we show an example of local explanations for a prediction.

\section{Data Augmentation}
For the CUB-200-2011 dataset, we first augment the dataset following ProtoPNet \citep{protopnet} pipeline\footnote{\url{https://github.com/cfchen-duke/ProtoPNet}} to have a fair comparison with ICICLE, then at runtime we use the data augmentation of PIP-Net \citep{pipnet}: we use TrivialAugment \citep{muller2021trivialaugment} twice. TrivialAugment applies a single random transformation per image. The first application is for location-based transformations (shear, rotation, translation), the second for color-based ones (brightness, sharpness, hue, contrast). For the Stanford Cars dataset, only PIP-Net's augmentation pipeline is applied.

\begin{table*}[h]
\centering
\begin{tabular}{l c c c p{5cm} }
\hline
\textbf{Phase} & \textbf{Model layers} & \textbf{Learning rate} & \textbf{Scheduler} & \textbf{Duration} \\
\hline
\multirow{1}{*}{Pretraining} 
& Backbone & $1 \cdot 10^{-4}$ & Cosine Annealing & \{10, 10, 10\} epochs on CUB, \newline \{10, 7, 5\} epochs on CARS \\
\hline
\multirow{3}{*}{Training} & Backbone & $1 \cdot 10^{-4}$ & \multirow{2}{*}{Cosine annealing} & \multirow{3}{5cm}{\{20, 15, 10\} epochs on CUB,\newline \{20, 13, 10\} epochs on CARS} \\
& $\tau$ & $1 \cdot 10^{-2}$ & &  \\
& Classification heads & $1 \cdot 10^{-3}$ & Cosine annealing with warm restart &  \\
\hline
\end{tabular}
\caption{CIP-Net's hyperparameters}
\label{tab:hyperparams}
\end{table*}

\begin{table*}[h]
\centering
\setlength{\tabcolsep}{6pt}
\begin{tabular}{|c|c|c|c|c|c|c|}
\hline
\textbf{Method} & \multicolumn{3}{c|}{\textbf{Final Avg. Task-Incremental Accuracy}} & \multicolumn{3}{c|}{\textbf{Final Avg. Class-Incremental Accuracy}} \\
\cline{2-7}
& 4 Tasks & 7 Tasks & 14 Tasks & 4 Tasks & 7 Tasks & 14 Tasks \\
\hline
EWC          & 0.456 $\pm$ 0.021 & 0.315 $\pm$ 0.037 & 0.287 $\pm$ 0.041 & 0.258 $\pm$ 0.019 & 0.152 $\pm$ 0.022 & 0.011 $\pm$ 0.009 \\
LWM          & 0.459 $\pm$ 0.072 & 0.416 $\pm$ 0.048 & 0.305 $\pm$ 0.022 & 0.233 $\pm$ 0.026 & 0.171 $\pm$ 0.016 & 0.080 $\pm$ 0.008 \\
LWF          & 0.375 $\pm$ 0.021 & 0.356 $\pm$ 0.024 & 0.250 $\pm$ 0.020 & 0.230 $\pm$ 0.011 & 0.171 $\pm$ 0.005 & 0.092 $\pm$ 0.008 \\
ICICLE      & \underline{0.654 $\pm$ 0.014} & \underline{0.645 $\pm$ 0.003} & \underline{0.583 $\pm$ 0.048} & \underline{0.335 $\pm$ 0.005} & \underline{0.203 $\pm$ 0.010} & \underline{0.116 $\pm$ 0.018} \\
\textbf{CIP-Net}      & \textbf{0.862 $\pm$ 0.008} & \textbf{0.921 $\pm$ 0.007} & \textbf{0.925 $\pm$ 0.012} & \textbf{0.589 $\pm$ 0.006} & \textbf{0.585 $\pm$ 0.009} & \textbf{0.389 $\pm$ 0.003} \\
\hline
PIP-Net C   & \multicolumn{6}{c|}{0.882 $\pm$ 0.005} \\
\hline
\end{tabular}
\caption{Final average (3 seeds) accuracy comparison for different numbers of tasks on Stanford Cars, demonstrating the
negative impact of the high number of tasks to be learned on models’ performance. Despite this trend, CIP-Net outperforms the baseline methods across all task numbers. At last, the PIP-Net (ConvNeXt) performance upper bound is presented.
Results of methods different from CIP-Net were taken from \citet{icicle}.}
\label{table:avg_acc_cars}
\end{table*}

\begin{table*}[h]
\centering
\footnotesize
\setlength{\tabcolsep}{1pt}
\begin{tabular}{|l|c|c|c|c|c|c|c|c|c|c|}
\hline
& \multicolumn{5}{|c|}{\textbf{Avg. Task-Incremental Accuracy}} & \multicolumn{5}{|c|}{\textbf{Avg. Class-Incremental Accuracy}} \\
\hline
\textbf{Method} & Task 1 & Task 2 & Task 3 & Task 4 & Avg & Task 1 & Task 2 & Task 3 & Task 4 & Avg\\
\hline
EWC & 0.24 $\pm$ 0.02 & 0.38 $\pm$ 0.07 & 0.54 $\pm$ 0.04 & 0.60 $\pm$ 0.05 & 0.44 $\pm$ 0.01 & 0.00 $\pm$ 0.00 & 0.06 $\pm$ 0.00 & 0.27 $\pm$ 0.03 & 0.53 $\pm$ 0.05 & 0.21 $\pm$ 0.01  \\
LWF & 0.17 $\pm$ 0.05 & 0.12 $\pm$ 0.01 & 0.23 $\pm$ 0.02 & 0.74 $\pm$ 0.06 & 0.30 $\pm$ 0.05 & 0.16 $\pm$ 0.03 & 0.00 $\pm$ 0.00 & 0.02 $\pm$ 0.00 & 0.54 $\pm$ 0.14 & 0.22 $\pm$ 0.02 \\
LWM & 0.19 $\pm$ 0.01 & 0.41 $\pm$ 0.01 & 0.43 $\pm$ 0.03 & \underline{0.77 $\pm$ 0.01} & 0.45 $\pm$ 0.02 & 0.03 $\pm$ 0.02 & 0.02 $\pm$ 0.02 & 0.08 $\pm$ 0.01 & \textbf{0.77 $\pm$ 0.04} & 0.18 $\pm$ 0.03 \\
ICICLE & 0.52 $\pm$ 0.02 & \textbf{0.66 $\pm$ 0.05} & \underline{0.71 $\pm$ 0.04} & 0.72 $\pm$ 0.00 & 0.65 $\pm$ 0.01 & 0.23 $\pm$ 0.01 & \underline{0.36 $\pm$ 0.02} & 0.31 $\pm$ 0.01 & 0.49 $\pm$ 0.02 & 0.35 $\pm$ 0.05 \\
CIP-Net (R) & \underline{0.74 $\pm$ 0.01} & 0.58 $\pm$ 0.11 & 0.66 $\pm$ 0.02 & 0.71 $\pm$ 0.07 & \underline{0.67 $\pm$ 0.01} & \underline{0.24 $\pm$ 0.01} & 0.25 $\pm$ 0.03 & \underline{0.37 $\pm$ 0.07} & 0.61 $\pm$ 0.06 & \underline{0.37 $\pm$ 0.01} \\
\textbf{CIP-Net (C)}      
& \textbf{0.83 $\pm$ 0.00} & \underline{0.63 $\pm$ 0.16} & \textbf{0.81 $\pm$ 0.04} & \textbf{0.81 $\pm$ 0.07} & \textbf{0.77 $\pm$ 0.01} & \textbf{0.25 $\pm$ 0.05} & \textbf{0.42 $\pm$ 0.02} & \textbf{0.51 $\pm$ 0.04} & \underline{0.70 $\pm$ 0.07} & \textbf{0.47 $\pm$ 0.01} \\
\hline
\end{tabular}
\caption{Average (3 seeds) task- and class-incremental accuracy across 4 tasks for each method on CUB-200-2011. (R) and (C) represent ResNet34 and ConvNeXt-tiny backbones respectively \citep{resnet34,convnexttiny}. Results of methods different from CIP-Net were taken from \citet{icicle}.}
\label{table:avg_acc_4ts}
\end{table*}

